\documentclass[letterpaper, 10 pt, journal, twoside]{IEEEtran} %

\IEEEoverridecommandlockouts                              %

\usepackage{graphicx} %
    \graphicspath{figures}
    \DeclareGraphicsExtensions{.pdf,.jpeg,.png}
\usepackage{textcomp, gensymb} %

\usepackage{amsmath,amsfonts, amssymb}
\usepackage{array}
\usepackage{textcomp}
\usepackage{stfloats}
\usepackage{url}
\usepackage{verbatim}
\usepackage{cite}
\usepackage{tabularx}
\usepackage{multirow}
\usepackage[linesnumbered,ruled,vlined]{algorithm2e}
\usepackage{algpseudocode}
\usepackage{rotating}
\usepackage{multirow}
\usepackage{lscape}
\usepackage{booktabs}
\usepackage{float}
\usepackage{xspace}
\usepackage[normalem]{ulem}
\usepackage{siunitx}
\usepackage{adjustbox}
\usepackage[dvipsnames]{xcolor}

\usepackage{pgfplots}
\pgfplotsset{compat=1.18}
\usepackage{xstring}
\usepackage{soul}       %

\usepackage[pagebackref,breaklinks,colorlinks=true, allcolors=blue]{hyperref}
\def\BibTeX{{\rm B\kern-.05em{\sc i\kern-.025em b}\kern-.08em
		T\kern-.1667em\lower.7ex\hbox{E}\kern-.125emX}}

\usepackage[nameinlink]{cleveref} %
\crefname{figure}{fig.}{Fig.} %
\Crefname{figure}{Fig.}{Fig.}

\title{
Event Spectroscopy: Event-based Multispectral and Depth Sensing using Structured Light
} %

\author{
Christian~Geckeler$^{*1,2}$,
Niklas~Neugebauer$^{*1,2,3}$,
Manasi~Muglikar$^{3}$,
Davide Scaramuzza$^{3}$,
and Stefano~Mintchev$^{1,2}$%
	\thanks{Manuscript received: August, 25, 2025; Revised November, 17, 2025; Accepted December, 6, 2025.}%
	\thanks{This paper was recommended for publication by Editor Soon-Jo Chung upon evaluation of the Associate Editor and Reviewers' comments.
		This work was supported by the Swiss National Science Foundation through the Eccellenza Grant number 186865, SONY R\&D Center Europe and the National Centre of Competence in Research (NCCR) Robotics.} %
\thanks{$^{*}$ These authors contributed equally to this work.}
\thanks{$^{1}$ Environmental Robotics Laboratory, Dep. of Environmental Systems Science, ETH Zurich, 8092 Zurich, Switzerland}
\thanks{$^{2}$Swiss Federal Institute for Forest, Snow and Landscape Research (WSL), 8903 Birmensdorf, Switzerland.}%
\thanks{$^{3}$ Robotics and Perception Group, University of Zurich, Switzerland.}%
\thanks{Corresponding author: {\tt\footnotesize cgeckeler@ethz.ch}}
\thanks{Digital Object Identifier (DOI): see top of this page.}
}

\graphicspath{ {./images/} }

\begin{document}
\bstctlcite{BSTcontrol}

\newcommand{\imageTable}[9][]{
\begin{tikzpicture}[node distance =0pt and 0.5cm]

\def\dirName{#2}
\def\separator{#4}

\def\sensorNames{#5}
\def\sceneNames{#3}
\def\sceneTitles{#6}
\def\sensorTitles{#7}

\StrCount{\sceneNames}{,}[\maxRow]
\StrCount{\sensorNames}{,}[\maxCol]
\pgfmathsetmacro{\nRows}{\maxRow+1}
\pgfmathsetmacro{\nCols}{\maxCol+1}

\def\vSpacing{#8}
\def \hSpacing{#9}

\pgfmathsetmacro{\imageHeight}{.8 * \vSpacing}

\pgfmathsetmacro{\rowStart}{0}
\pgfmathsetmacro{\rowEnd}{\nRows-1}
\def\colStart{1}
\pgfmathsetmacro{\colEnd}{\maxCol + 1}

  \foreach \scene [count=\y from \rowStart] [evaluate=\y as \yp using \vSpacing*\y] in \sceneNames {
    \foreach \sensor [count=\x from \colStart] [evaluate=\x as \xp using \hSpacing*\x] in \sensorNames {
    \ifx\\#1\\
    \else
        \pgfmathsetmacro{\xp}{\hSpacing*\y}
        \pgfmathsetmacro{\yp}{\vSpacing*\x} 
    \fi
    \node (\x\y) at (\xp,\yp){\includegraphics[height=\imageHeight cm]{
       \dirName/\scene\separator\sensor}};
       }
    }
  \foreach \sensor [count=\x from \colStart] [evaluate=\x as \xp using \hSpacing*\x] in \sensorTitles {
  \ifx\\#1\\
        \pgfmathsetmacro{\yp}{\vSpacing*(\nRows-0.5) + (0.05*\baselineskip)} 
  \else
        \pgfmathsetmacro{\yp}{\vSpacing*\x}
        \pgfmathsetmacro{\xp}{-\hSpacing}
  \fi
  \node (\x)[text width=\hSpacing cm,align=center] at (\xp,\yp) {\sensor};
  }
  \foreach \scene [count=\y from \rowStart] [evaluate=\y as \yp using \vSpacing*\y] in \sceneTitles {
  \ifx\\#1\\
        \pgfmathsetmacro{\xp}{0} 
  \else
        \pgfmathsetmacro{\xp}{\hSpacing*\y}
        \pgfmathsetmacro{\yp}{\vSpacing*\nCols+1}
  \fi
    \node (\y) at (\xp, \yp) {\scene};
  }
\end{tikzpicture}
}

\maketitle

\markboth{IEEE Robotics and Automation Letters. Preprint Version. 01, 2026}
{Geckeler \MakeLowercase{\textit{et al.}}: Event Spectroscopy} 

\begin{abstract}

Uncrewed aerial vehicles (UAVs) are increasingly deployed in forest environments for tasks such as environmental monitoring and search and rescue, which require safe navigation through dense foliage and precise data collection. 
Traditional sensing approaches, including passive multispectral and RGB imaging, suffer from latency, poor depth resolution, and strong dependence on ambient light—especially under forest canopies. 
In this work, we present a novel event spectroscopy system that simultaneously enables high-resolution, low-latency depth reconstruction with integrated multispectral imaging using a single sensor.
Depth is reconstructed using structured light, and by modulating the wavelength of the projected structured light, our system captures spectral information in controlled bands between $650$ nm and $850$ nm.
We demonstrate up to 60\% improvement in RMSE over commercial depth sensors and validate the spectral accuracy against a reference spectrometer and commercial multispectral cameras, demonstrating comparable performance. 
A portable version limited to RGB is used to collect real-world depth and spectral data from a Masoala Rainforest. %
We demonstrate color image reconstruction and material differentiation between leaves and branches using this spectral and depth data. 
Our results show that adding depth (available at no extra effort with our setup) to material differentiation improves the accuracy by over 30\% compared to color-only method.
Our system, tested in both lab and real-world rainforest environments, shows strong performance in depth estimation, RGB reconstruction, and material differentiation—paving the way for lightweight, integrated, and robust UAV perception and data collection in complex natural environments.
\end{abstract}


\section{Introduction}
\label{sec:intro}

\begin{figure}[t]
    \includegraphics[width=0.5\textwidth]{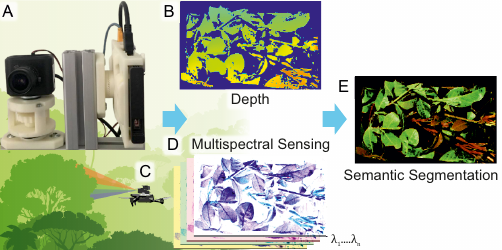}
    \caption{\textbf{Event Spectroscopy}: We propose an all-in-one solution for depth sensing, color image reconstruction and multispectral sensing. 
    Usable for instance, for an uncrewed aerial vehicle (UAV) navigating in forest environments (C).  A) Our portable setup consisting of an event camera and a projector as an illumination source 
    B) Generated depth of scene acquired using structured light,
    D) spectral image reconstructed using events, 
    E) material segmentation of leaves using spectral and depth data.} 
    \label{fig:eye} 
    \vspace{-1.5em}
\end{figure}

\IEEEPARstart{T}{here} is an increased demand for uncrewed aerial vehicles (UAVs) flying in forests \cite{Loquercio2021, Ren2025}, both for environmental monitoring \cite{Geckeler2025d, Charron2020}, and for search and rescue applications \cite{Schedl2021}.
Vision is essential for obstacle sensing and data collection to enable these applications. For instance, environmental monitoring tasks such as sensor deployment \cite{Geckeler2022,  Geckeler2023b} or sample collection \cite{Aucone2023a, Charron2020, Kirchgeorg2024} require UAVs to fly in close proximity to tree branches and foliage, where fast-moving, thin structures must be detected under dynamic and often challenging lighting conditions. %
Flying in close proximity to trees makes interaction with foliage probable, and while it has been shown that UAVs can push aside flexible twigs and leaves while avoiding thicker branches\cite{Aucone2024}, proper sensing to differentiate between branches and foliage is needed\cite{Geckeler2024}. This requires not only low latency and high resolution depth sensing of thin and fine structures, but also accurate differentiation between woody branches and foliage, for instance using multispectral sensing. %

Vision is also used for data collection tasks, including multispectral imagery for assessing tree health, physiological traits, and species identification \cite{Xu2020a}. Current methods typically use passive spectral sensors mounted on satellites, aircraft, or UAVs operating far above the forest canopy \cite{Jarocinska2023}. To generate more informative insights, higher spatial resolution data is required, which can be achieved by capturing images closer to or from within the canopy. However, these sensors are highly dependent on ambient illumination, and the pronounced variations in lighting intensity below the canopy make the use of standard multispectral cameras extremely challenging. Even setting aside this limitation, a UAV would currently still need to carry multiple conventional sensors; a high resolution depth camera for obstacle sensing, an RGB camera for scene video and context, and a multispectral camera to capture multispectral data.

In this work, we propose a single event-based structured light solution which can simultaneously deliver high-resolution and low latency depth reconstruction with integrated multispectral sensing. 
Based on previous work, the integrated system utilizes an event camera with structured light to reconstruct high-fidelity depth with low latency. By changing the wavelength of the light used for structured light we can not only reconstruct RGB images, but also perform multispectral sensing by projecting the desired wavelength - with lower latency than available multispectral sensors, using our novel method. To the best of our knowledge, this is the first integrated event-based system which provides both depth as well as multispectral sensing. First, we validate the quality of our depth reconstruction by demonstrating up to 60\% average improvement in the RMSE over other commercial off-the-shelf depth sensors. Next, we demonstrate event-based multispectral sensing using wavelengths between 650 nm to 850 nm in a lab setting and compare our results to a commercial multispectral sensor and a ground truth spectrometer. Furthermore, we demonstrate RGB color image reconstruction using a portable version limited to RGB light projection, as well as material differentiation utilizing both the depth and spectral data on real-world data from a Masoala Rainforest.
\vspace{-0.5em}

\section{Related Works}

Our solution provides event-based multispectral and depth sensing. This requires both spectral imaging as well as event-based luminance recovery and geometry estimation. 
The following sections present current solutions to these problems as well as our methods.

\subsection{Spectral Imaging}

Spectral imaging captures images in multiple wavelength bands, with established, widespread applications.
The resulting $(X, Y, \lambda )$ data cube contains both spatial $(X, Y)$ information and spectral information, $\lambda$, per wavelength.
Data capture methods can be divided into two main groups: scanning and snapshot methods.
Snapshot methods capture the full data cube during a single integration period of the sensor, whereas scanning methods achieve the same over multiple periods.
Among scanning methods, tuned filters represent a popular option. These include a rotating filter wheel, an electro-mechanical Fabry-Ferot filter \cite{MEMSFabryFerot}, liquid-crystals \cite{Gupta2008}, or acousto-optic tunable filters \cite{poger2001multispectral}.
\newline
Multiplexing techniques such as Fourier Transform Spectroscopy\cite{descour1996throughput} and Computed Tomography Multi-Spectral-Imaging\cite{Mooney1995} allow for the reconstruction of multiple wavelengths from fewer images than the number of wavelengths at the expense of
some artifacts.
\newline
Classic snapshot methods make use of multiple sensors in the form of Bayer Pattern filter layouts, beam-splitters, or simply by using multiple full camera sensors.
\newline
Recent developments in compressive sensing like CASSI \cite{Gehm2007} use a two-dimensional patterned grating to reconstruct all images of all wavelengths in a single shot. While the computational complexity required to decode the image presents a major bottleneck, improvements have made this method viable for high-resolution imaging in both spatial and spectral dimensions \cite{Wang17PAMI}.
\newline
In this work, we employ an inverse variant of a tuned filter. Instead of filtering the light entering the imaging sensor, we illuminate the scene with light of a specific wavelength band, and measure the resulting change in reflected light off the scene with respect to ambient illumination. While this active approach requires more power and additional components compared to passive alternatives such as color filter arrays or learned methods, it enables simultaneous capture of both high fidelity spectral and depth information, and is more robust in variable and low-light conditions. Particularly for UAV applications, having an integrated system capable of providing depth, multispectral and RGB sensing as opposed to  separate devices for each, compensates the increased power consumption of the active approach when considering payload restrictions.

\subsection{Event Luminance Recovery and Geometry Estimation}
As events naturally compress visual information, estimating the absolute intensity solely from events is a challenging task.
Prior methods demonstrate reconstruction of a grayscale image up to an unknown intensity value through the use of data-driven priors \cite{Rebecq2018}.
Color intensity can also be recovered by using color event cameras \cite{Scheerlinck2019}, or through setups with multiple cameras and appropriate color filters \cite{Marcireau2018}. Multispectral sensing for face recognition through the addition of a single infrared filter was shown in \cite{Himmi2024}.
In \cite{Cao2024}, the illumination-dependent noise characteristics of event cameras are used to reconstruct the intensity, however, mainly for static scenes where there is no relative motion between the camera and the scene.

Recent progress in combining active illumination with event cameras \cite{Matsuda15ICCP, ESL, Muglikar2021a}, has resulted in accurate and high-speed geometry estimation.
The setup consists of a laser scanning projector with an event camera in a stereo camera configuration. The projector illuminates the scene with a known pattern and the event camera observes the reflection of this pattern, which is then used to triangulate the depth of the scene.
While it was shown \cite{Matsuda15ICCP} that the events generated by this reflection are independent of scene reflectivity, this is only true for an ideal sensor.
The second order noise characteristics of an event camera, however, are illumination dependent thus creating non-idealities.
This principle was used in \cite{Ehsan2022} to reconstruct the absolute intensity by observing how many events were generated through the reflected light.
For darker objects, the event count would be lower as the incoming light has a lower intensity, whereas brighter materials reflect more light and thus trigger more events.

Several approaches for recovering only spectral information or only depth using event cameras have been proposed. For spectral information these include using the chromatic aberrations from a ball lens while physically adjusting the focal length of the event camera \cite{Arja2025}, to using a diffraction grating and rotating sweeping mirror to capture spectral information \cite{Yu2025}. Looking only at depth, previous work has shown this can be reconstructed using point-spread function engineering  \cite{Shah2024a} or coded apertures \cite{Habuchi2024}, but is missing spectral information. The utility of event cameras for downstream tasks such as semantic segmentation \cite{Alonso2019} have also been demonstrated. While non-event based, conventional systems can capture hyperspectral and depth data with a single camera using learned methods \cite{Baek2021}, the resulting depth has low resolution and fidelity.

In this letter, we expand on previous work \cite{Muglikar2021}, utilizing the structured light projector not only for depth reconstruction, but also for multispectral sensing. In contrast to previous approaches, by extracting both depth and the spectral properties of scenes in conjunction with the event camera, a unified, high-resolution, low-latency system is created, providing rich scene data for downstream tasks.

\section{Event-based Multispectral and Depth Sensing}
This section provides an overview of the system and introduces the basics of an event camera (\Cref{sec:eventcam}) and multispectral imaging with an event camera (\Cref{sec:emi}).

\subsection{System Overview}

Our system captures both high-resolution depth and multispectral reflectivity using an event camera and a projector as a stereo pair. Depth is reconstructed using an event-based structured light method \cite{Muglikar2021a}. To generate depth, a point scanning laser projector shortly illuminates one pixel at a time, scanning the full projected image in the process. Due to the microsecond temporal resolution of event cameras, every pixel illumination generates an event and is matched to the origin in the projected image. From this, the depth of all illuminated pixels is then computed using standard stereo geometry.
To estimate spectral reflectivity, we observe the repeated light flashes of the projector while progressively increasing the contrast threshold of the camera (reducing sensitivity). For every pixel, the minimum sensitivity at which an event still occurs indicates its relative reflectivity at that wavelength. Repeating this procedure for each spectral channel yields a multispectral image that is aligned with the depth map.

\subsection{Event Camera}
\label{sec:eventcam}
Event-cameras are bio-inspired sensors that asynchronously measure \emph{changes} (i.e., temporal contrast) in illumination at every pixel, at the time they occur \cite{Lichtsteiner2008,Suh2020,Finateu20isscc, Posch11ssc}.
In particular, an event camera generates an event $e_k = (\mathbf{x}_k,t_k,p_k)$ at time $t_k$ when the difference of logarithmic brightness at the same pixel $\mathbf{x}_k=(x_k,y_k)^\top$  reaches a contrast threshold $C$:
\begin{equation}
\label{eq:egm}
    L(\mathbf{x_k},t_k) - L(\mathbf{x_k},t_k-\Delta t_k) = p_k C,
\end{equation}
where $p_k \in \{-1,+1\}$ is the sign (or polarity) of the brightness change, 
and $\Delta t_k$ is the time since the last event at the pixel $\mathbf{x}_k$. The result is a sparse sequence of events which are asynchronously triggered by illumination changes.

\subsection{Active Illumination with  Event Cameras}
Significant literature on event-based vision has assumed constant brightness, where events are only generated by relative motion between the camera and the scene \cite{Gallego2019}.
However, the presence of an active illumination source changes the event generation model.

In \cite{Matsuda15ICCP}, it was shown that the change in illumination ($\delta I$) measured by an event camera, when a projector with intensity $I_p$ illuminates a point of reflectivity $T$ under uniform ambient light $I_a$, is given by:
\begin{equation}
    \delta I = log (\frac{I_p + I_a}{I_a})\\
             =  log ((I_p + I_a) \times T) - log(I_a \times T) \\
\end{equation}
As evident from this equation, event generation is independent of the scene reflectivity $T$.
This allows for improved depth reconstruction for scenes with varying reflectivity \cite{Muglikar2021}.
However, this equation also suggests that it is infeasible to recover scene illumination from these sensor measurements.
In the next section, we show how it is nonetheless possible to recover relative scene reflectivity in this setup.

\subsection{Event-based Spectral Imaging}
\label{sec:emi}
In an ideal sensor, the measured change in intensity due to projector illumination is independent of the scene reflectivity \cite{Matsuda15ICCP, Muglikar2021}.
However, the noise and sensor characteristics of the event camera have an illumination dependence which we exploit to capture the scene reflectivity.

\begin{figure}
    \centering
    \includegraphics[width=\columnwidth]{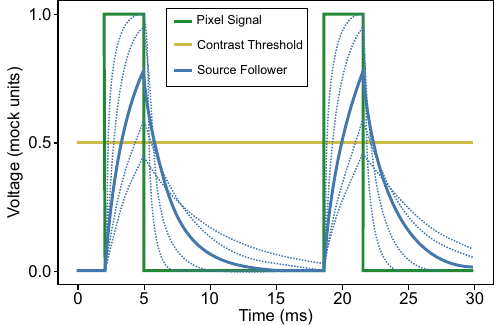}
    \caption[Qualitative Simulation of the Source Follower with an Oscillating Signal]{The simulation demonstrates how a change in the source follower bandwidth can regulate the observed signal amplitude downstream. Responses for different follower gains are shown as dotted lines.}
    \label{fig:sourceFollower}
\end{figure}

In a typical event camera pixel, the photocurrent generated by the photodiode is amplified and then processed by a source follower circuit, which attempts to match its output voltage to the incoming signal. 
The speed at which the source follower can track changes in the signal depends on both its bias current and the amount of incoming light. 
Consequently, pixels exposed to brighter light respond more quickly, while those in darker regions respond more slowly \cite{DVSBiases2023}.
\Cref{fig:sourceFollower} showcases the effect of the source follower bandwidth on the amplitude of the measured signal.
This relationship between the light intensity and the source-follower voltage introduces a dependence on the scene reflectivity.
Our method uses the fact that the bandwidth of the source follower in each pixel depends on the absolute intensity that its photoreceptor receives.
As a result, when the illumination changes rapidly, pixels corresponding to darker or less reflective regions may not respond quickly enough to reach the event-generation threshold, even if the relative change in light intensity is significant. 
This means that only pixels with reflectivity above a certain threshold will generate events, while those below this threshold will not. By carefully varying the threshold and observing which pixels generate events, we can estimate a lower bound for the reflectivity at each pixel. This method allows us to exploit the sensor's non-idealities to recover information about the relative reflectivity of the scene.

\section{Characterization}
This section evaluates the performance of our event-based depth and multispectral imaging system.
We begin by describing the hardware setup and the evaluation baselines and performance metrics.
We evaluate the performance of our system for depth estimation, comparing both qualitative and quantitative results to other depth sensors.
We then evaluate the performance of our system for multispectral imaging, comparing both qualitative and quantitative results with those of other multispectral imaging systems.
Finally, we evaluate the capability of our system for two tasks: material differentiation using a full-spectrum light source and material segmentation using a limited-spectrum projector and depth.
To the best of our knowledge, there exists no dataset on which the proposed approach can be evaluated.
Therefore, we collect our own dataset by first building our prototype system using an event camera and a projector (\Cref{fig:eye}A).

\subsection{Hardware Setup}

We use a Prophesee Gen3 event camera \cite{Posch11ssc} with a resolution of $640\times480$ pixels.
This sensor provides regular events without exposure measurements; these events are then used for depth estimation and spectral imaging.
A lens with a field of view of $60^\circ$ is used throughout all experiments.
The camera offers two parameters to adjust: the contrast threshold and the source follower gain. Both have the same (though inverted) downstream effect on event generation. Empirically, the working range was determined to lie between 1550 and 1800 for the contrast threshold and 415 to 600 for the source follower gain. Thus, we have a slightly higher resolution stepping of 185 for the source follow vs. 150 for the contrast threshold.

\begin{figure}
    \centering
    \includegraphics[width=0.5\textwidth]{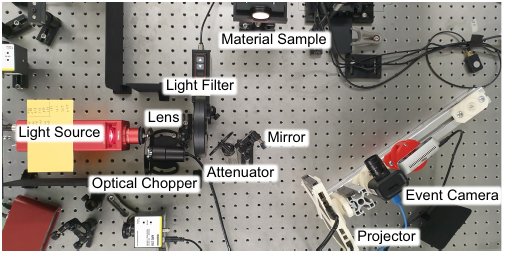}
    \caption{The full spectrum lab illumination setup with our event-camera setup.}
    \label{fig:generalSetup}
    \vspace{-1.5em}
\end{figure}

Two projector setups are used in the experiments, a full-spectrum illumination setup and a portable setup.

The portable setup employs a Sony  MP-CL1A projector  with a resolution of $1920\times720$ pixels, capable of projecting three distinct wavelengths (red, green, blue). This setup is used for data capture for color image reconstruction and the outdoor experiments. The generation of a full depth image takes slightly less than 16 ms, and around one second to capture spectral information across the full image for one wavelength. The quality and maximum range of the depth and spectral sensing depend on the power and resolution of the projector. Most of the following scenes were captured within 1 m of the camera.
For multispectral analysis, a custom multispectral projector is built (Fig. \ref{fig:generalSetup}). This setup is used to compare the spectral response of our setup to a ground-truth spectrometer and a commercial multispectral camera. It consists of a full spectrum light source and six filters corresponding to wavelength bands of 10 nm each.
It is combined with an optical chopper that flickers the illumination at a frequency of $100 Hz$. Although highly accurate and fully customizable to any desired wavelength, this setup is too large to be used outside of the lab setting.

\subsection{Evaluation}
The performance of the system is evaluated for depth estimation and color image reconstruction, using the portable setup (\Cref{fig:eye}A).

\subsubsection{Depth estimation}
\label{sec:eval:depth}
For depth estimation, we compare the performance of our system to the existing depth sensors Microsoft Kinect V2 (Kinect), PMD Pico-Flexx 2 (Pico) \cite{pmdtec}, and Intel RealSense D435 \cite{IntelRealSense}. We evaluate the RealSense camera in both the Structured Light (RealSense) and Direct Stereo mode (RS Stereo).
The ground truth point-cloud is collected using a FARO Focus 3D S 120 laser scanner at a resolution of 58 points per degree horizontally and 33 points per degree vertically.
The point-clouds generated by these depth sensors are aligned to the ground truth point-cloud using ICP.
We compare the sensor point-clouds and the ground-truth point-cloud using the root-mean square error (RMSE) and the Chamfer distance.
RMSE computes the distance between corresponding points in the point-cloud and ground-truth point-cloud. The Chamfer distance, instead, compares each point in the point-cloud to the nearest neighbor in the ground-truth point-cloud. Thus, when there are large discrepancies between the point clouds, the Chamfer distance provides a more informative metric.

\begin{figure}
    \centering
    \includegraphics[width=.5\textwidth]{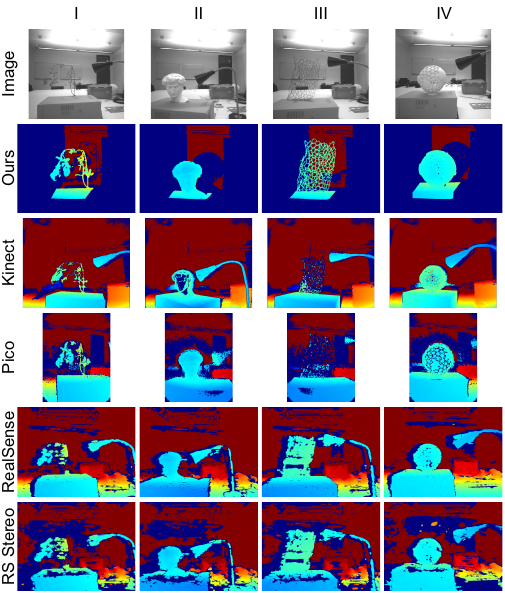}
    \caption{Comparing depth accuracy of different sensors when imaging challenging scenes such as thin and hollow structures or strong reflections.}
    \label{fig:depth_comparison}
\end{figure}

\begin{table*}[t]
    \centering
    \begin{adjustbox}{max width=\linewidth}
    \setlength{\tabcolsep}{4pt}
    {\small
        \begin{tabular}{rrrrrrrrrrr}
        \toprule
        scene  &  \multicolumn{2}{c}{Ours}  &  \multicolumn{2}{c}{Kinect}  &   \multicolumn{2}{c}{Pico}  &  \multicolumn{2}{c}{RealSense}  &   \multicolumn{2}{c}{RS Stereo}  \\
         Metrics   &  RMSE $\downarrow$ & Chamfer $\downarrow$ &  RMSE $\downarrow$ & Chamfer $\downarrow$ &  RMSE $\downarrow$ & Chamfer $\downarrow$ &  RMSE $\downarrow$ & Chamfer $\downarrow$ &  RMSE $\downarrow$ & Chamfer $\downarrow$ \\
        \midrule
            Branch &  \textbf{0.544} &\textbf{0.819} &   0.751& 0.885 & 0.947 & 1.121 &             0.903 &  1.848&             1.203 &             1.722 \\
            Buddha & 0.335 & \textbf{0.363} & 1.254 & 0.902  & 1.370 & 0.979 &    \textbf{0.226} & 0.381  & \underline{0.325} &0.434 \\
            David &    \textbf{0.295} & \textbf{0.525} & 1.669 & 1.357 & 1.453 & 1.257 & 0.422& 0.789 & 0.635 & 0.820 \\
            Globe &\underline{0.269}& \textbf{0.591} & 0.807 & 0.713 & 0.771 & 1.282 & \textbf{0.26} & 0.797 & 0.331 & 0.823 \\
            Lamp &\underline{0.817} &  \textbf{0.982} &   1.480 &  1.519& 0.929  & 1.494 & \textbf{0.481} & \underline{1.01}  & 1.073 & 1.203 \\
        \bottomrule
        \end{tabular}    
    }
    \end{adjustbox}
\caption{Comparison of depth sensors over all reconstructed scenes using Chamfer distance (cms) and RMSE (cms). Lower is better}
\label{tab:depthTable}
\vspace{-1.5em}
\end{table*}

We show qualitative results on depth reconstruction of all depth sensors in \Cref{fig:depth_comparison}.
The RealSense depth camera (RealSense) uses active stereo depth estimation, which projects a sparse random dot pattern.
The sparse depth is filled using a hole filling approach, resulting in an overestimation of the depth and missing structural details.
The Kinect, on the other hand, uses time-of-flight based depth sensing, where each pixel has a pulsed light emitter and detector. 
This makes the pixel size much larger; therefore, despite the densely illuminated scene, the sensor fails to capture thin structures such as the wire frame (\Cref{fig:depth_comparison}, Column III).
Moreover, due to inherent limitations with time-of-flight sensors, it cannot capture the depth of face of statue of David (\Cref{fig:depth_comparison},  Column II) due to strong inter-reflections. Similarly for the time-of-flight Pico camera.
Our method, on the other hand, achieves a better balance by capturing the intricate details of the wireframe and the globe structure because of its dense projected pattern.
Quantitative comparisons for the standard illumination show up to 125\% improvement in the Chamfer distance for our event-based depth sensor when compared to state-of-the-art depth sensors for the same framerate, \Cref{tab:depthTable}. %

\paragraph{Effect of Illumination}
\begin{table}
    \centering
\begin{tabular}{rrrrrrr}
\toprule
Illumination & Ours & Kinect &  Pico & RealSense &  RS Stereo \\
\midrule
    Low Lux& \textbf{0.656} &   1.075 & 1.227 &            \underline{0.965} &             1.000 \\
    Bright Lux & \textbf{0.691} &   \underline{1.142} & 1.195 &            1.728 &             2.253 \\
\bottomrule
\end{tabular}
\caption{Effect of illumination on reconstructed error in cms. Lower is better}
\label{tab:depth_illum}
\vspace{-2em}
\end{table}

The effect of scene illumination on depth reconstruction can be seen in \Cref{tab:depth_illum}. First, depth is captured in a low-light scene, then bright external lighting is introduced. In general, performance is greatly reduced, with RS Stereo having a 56\% increase in error, and the RealSense a 44\% increase in error. Our method exhibits only a negligible performance drop of 5\%, resulting from overexposure of event frames. While the time-of-flight methods (Kinect, Pico) exhibit similar performance across both low and bright illumination (with the error of the Pico actually decreasing), the absolute error of our method is still by far the lowest, almost half of the Pico. This demonstrates that our method can produce high quality depth, maintaining performance across different lighting conditions, with improvements of at least 39\% and up to 226\% compared to other state-of-the-art frame-based depth sensors.

\subsubsection{Color Image Reconstruction}
\label{sec:eval:color}
In this section, we evaluate the spectral imaging capability of our setup.
We compare our method against the event counting baseline proposed in \cite{RGBD} both for color accuracy and image reconstruction accuracy.

\paragraph{Color Accuracy}
We compare the color accuracy on a printed version of an ISO $1233:2017$ conforming chart.
The chart features $16$ color blocks which are distributed over the sRGB spectrum.
We measure the mean color for each block and compare it with the ground-truth using $L2$ distance in the CIE $1976$.
This CIE is based on human perception of colors and serves as a standard reference for understanding and quantifying color.

We reconstruct color images by individually projecting only red, green, and blue light and capturing their images separately.

\begin{table}[!ht]
    \centering
    \setlength{\tabcolsep}{3pt}
    \begin{tabular}{c|c|c|c}
        \toprule
        Method & RMSE & RMSE (wb) & RMSE (curve) \\
        \hline
        Ours (PR Bias, 5100 ms) & \textbf{21.4938} & \textbf{19.8148} & \textbf{16.6109}  \\
        Event Counting\cite{RGBD} (6000 ms) & 33.1656 & 31.6368 & 26.0675 \\
    \bottomrule
    \end{tabular}
    \caption[Test Chart Color Errors]{Mean error between captured color values and their ground-truth on the test chart. Lower is better.}
    \label{table:Color_Error}
    \vspace{-1.5em}
\end{table}

\Cref{table:Color_Error} shows the RMSE over all 16 color blocks for different image correction techniques.
We use the information from the grayscale blocks on the chart 
to apply these corrections.
After white-balancing (wb) and linearization of the images (curve), the color accuracy is greatly improved. This can be attributed mainly to the higher dynamic range that our method exhibits.  For our method, we utilize the photoreceptor gain (PR Bias) to change the threshold for event generation.
In all scenarios, our method greatly outperforms the Event Counting method, with improvements in RMSE above 50\%.

\paragraph{Image Reconstruction Accuracy}
\begin{figure}
    \centering
    \includegraphics[width=\columnwidth]{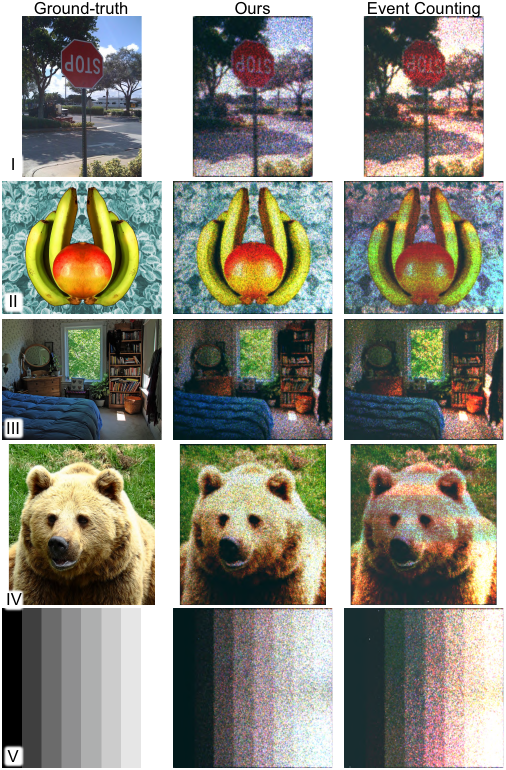}
    \caption[Captures of Projected Images]{Samples of reconstructed color-corrected images with ground-truth (left), ours (center), and using Event Counting (right)\cite{Ehsan2022},}
    \label{fig:coco_samples}
    \vspace{-1.5em}
\end{figure}

We also evaluate our approach to reconstruct more complex images, such as from the MSCOCO dataset \cite{MSCOCO}.
The results in \Cref{fig:coco_results} show a clear quantitative improvement in the RMSE between the reconstructed image pixels and ground-truth over the baseline method \cite{RGBD} by more than 24\%. In this case, we change the contrast threshold for event generation directly using the ON Bias (DIFF\_ON). Qualitative results can be seen in \Cref{fig:coco_samples}. This improvement can mainly be traced to two improvements: 
(i) Our method achieves a better dynamic range than the baseline, allowing it to resolve bright and dark regions more accurately. (e.g. \Cref{fig:coco_samples} V)
(ii) We are less affected by buffer overflow artifacts of the event camera because of its inherent redundancy and lower event-rate when capturing bright parts of the image.

\begin{table}[!htp]
    \centering

\begin{tabular}{lcccccc}
\toprule
& \multicolumn{3}{c}{Ours (ON Bias)} & \multicolumn{3}{c}{Event Counting\cite{RGBD}}\\
& raw & wb & curve & raw & wb & curve\\
\cmidrule(lr){2-4} \cmidrule(lr){5-7} 
red	&77.97 & 77.97 & \textbf{39.52} &\textbf{65.84} &\textbf{65.84} &51.90 \\
green &\textbf{34.06} &\textbf{ 45.48}& \textbf{33.22} &43.01 &46.57 &45.28 \\
blue & \textbf{41.12} & \textbf{39.63}& \textbf{36.56} &57.25 &45.62 &46.29 \\
\midrule
mean &\textbf{ 56.18} & 57.51& \textbf{36.88} &58.11 &\textbf{54.28} &48.51 \\
\bottomrule
\end{tabular}

    \caption[Image Capture Errors]{Root Mean Squared error between the captured images and their ground truth. Colors are in RGB color space.}
    \label{fig:coco_results}
    \vspace{-2em}
\end{table}

\section{Multispectral Segmentation}
In this section, we show the application of our setup for material segmentation.
We deploy our method using two setups : using the full spectrum light source for lab material identification (\Cref{sec:fullspectrum}), and using our portable setup for outdoor scenes (\Cref{sec:realforest}).

\subsection{Full Spectrum Indoor Scenes}
\label{sec:fullspectrum}

\begin{figure*}[!htp] %
    \centering
    \includegraphics[width=\linewidth]{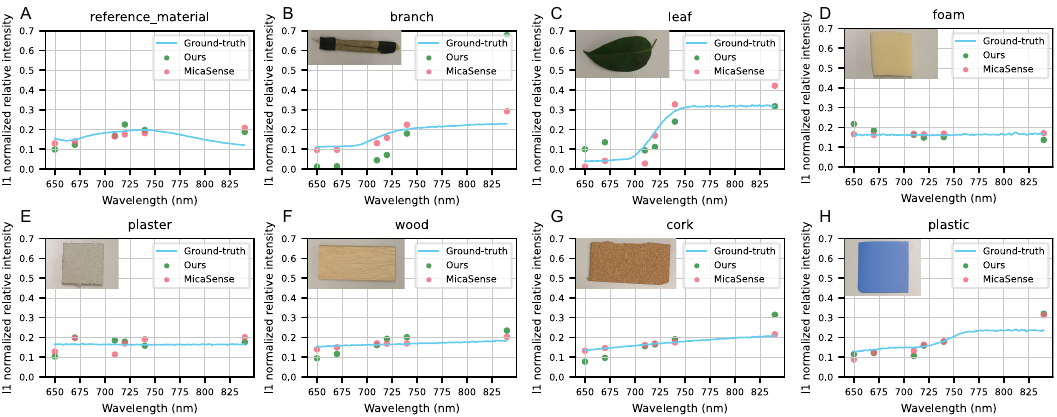}
    \caption{Material classification: spectral responses for different materials from a ground truth spectrometer (Ground-truth, blue), our proposed event-based system (Ours, green), and a commercial multispectral camera (MicaSense; MicaSense RedEdge MX Dual, pink). }
    \label{fig:materials}
    \vspace{-1em}
\end{figure*}

The material samples are illuminated using the full spectrum light source (\Cref{fig:generalSetup}), and the response is captured by the respective sensor. \Cref{fig:materials} shows the response curves for each sample as captured by the ground truth spectrometer, a commercial multispectral sensor, and our system. The materials are: a 99\% reflective reference panel (used for calibration), a branch, a leaf, foam, plaster, wood, cork, and plastic.
The commercial 10-band multispectral camera is the MicaSense RedEdge-MX Dual camera system, which serves as our baseline.
For all of the samples, the ground truth is collected using a spectrometer which most accurately measures the reflected light.
Qualitatively, the reflectance measured by the event camera follows the ground truth curves well. Beyond this, we see that there is comparable performance between the commercial multispectral MicaSense sensor, and our system, whereas our system tends to be closer to the ground-truth.
The normalized measurements are quite sensitive to outliers, which can cause particularly high errors, for instance for the branch, in which the reflectance for 850nm is far too high.
In general, we see that our performance can deliver comparable or even more accurate multispectral data when compared to a commercial multispectral sensor across a variety of different materials.

\subsection{Real Forest Material Differentiation Demonstration}
\label{sec:realforest}

Here, we demonstrate material segmentation of branches and leaves from real-world data, captured from the forested areas of the Masoala Rainforest in Zurich Zoo using the portable, $3$-wavelength illumination system (\Cref{fig:eye}A). 
A semi-supervised segmentation pipeline is used. First, segments are generated using a min-cut algorithm, where neighboring pixels are separated if they have low similarity, based on their RGB-D vector.
The segments are then classified with a shallow VGG-inspired Convolutional Neural Network with two convolutional and three fully connected layers.
For classifier training, $31$ RGB-D scenes are divided into $21$ training sequences and $10$ test sequences.
At the end, the pipeline predicts one of three labels for each pixel: leaves, branches or background.
\begin{table}[t]
    \centering
    \begin{tabular}{lrrr}
    \toprule
    channels  & \multicolumn{3}{c}{IoU}\\
     & leaves & branches & mean\\
    \midrule
    depth  & 0.534 & 0.045 & 0.365\\
    RGB  & 0.681 & 0.200 & 0.441\\
    RGBD  & \textbf{0.706} & \textbf{0.288} & \textbf{0.497}\\
    \bottomrule
    \end{tabular}
    \caption{IoU results for each of the classification labels when the pipeline is using only RGB, only depth, or both.}
    \label{tab:segmentation_results}
    \vspace{-2em}
\end{table}

\begin{figure}
    \centering
    \includegraphics[width=\columnwidth]{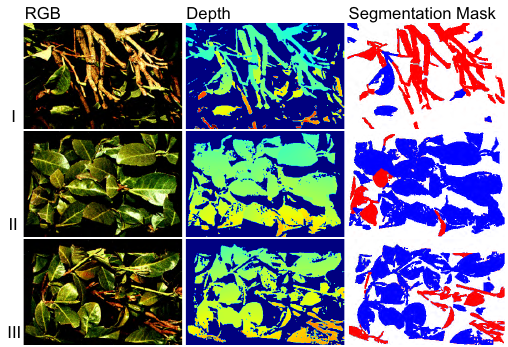}
    \caption{Three (I-III) sample segmentation results, with reconstructed RGB (A), depth (B), and computed segmentation mask (C), in red (branches) and blue (foliage).}
    \label{fig:segment}
    \vspace{-1.5em}
\end{figure}

We compare the performance of segmentation using only RGB images, only depth, and RGB-D fusion using the intersection-over-union (IoU) score for leaves and branches, these results are summarized in \Cref{tab:segmentation_results}. Qualitative results for the combined approach can be seen in \Cref{fig:segment}.
Since leaves are both more present and individually larger in most images, their class is generally easier to detect. Inter-class variation is also lower for leaves, and many errors are due to misclassification of brown or overexposed leaves as branches. 

Overall, the results suggest that RGB information is more important for the semantic segmentation task than only using depth images. 
The combination of the two, however, improves the segmentation results by 30\% (\Cref{tab:segmentation_results}) and yields the best results. This showcases the importance of diverse multi-modal data (both spectral data and depth) for downstream tasks, such as material differentiation.

\section{Conclusion}

This paper demonstrates the potential of event spectroscopy, generating high-resolution and low-latency depth reconstructions as well as multispectral sensing through an integrated event-based structured light system. 
We validate in lab conditions that, by varying the projected wavelength for structured light depth sensing, the event camera can effectively operate as a multispectral sensor, achieving performance comparable to that of conventional commercial off-the-shelf multispectral sensors.
This also enables accurate RGB color image reconstruction from events.
Finally, we demonstrate material differentiation in real-world forest data by leveraging both depth and spectral information, further highlighting the versatility and practical utility of our approach.

Our current prototype still suffers from two main limitations:
the portable prototype is limited to RGB multispectral sensing due to the availability of commercial projectors. 
Extending the system to support true multispectral sensing will require either a full-spectrum light source with appropriate wavelength filters or separate light sources, such as multispectral LEDs, for each desired wavelength. Since every wavelength is captured separately, capture time increases in proportion to the number of wavelengths. Reducing the spatial resolution and alternating wavelengths per pixel, or projecting complimentary, non-overlapping patterns to capture multiple wavelengths simultaneously could alleviate these issues.
Second, the projected illumination may be insufficient under strong ambient light or if the sensor is far away from the sample, particularly for outdoor applications.
Addressing this limitation will require higher output power or more efficient light sources to ensure reliable detection. Additional inputs, for instance, polarization cues, could also improve the material classification.
Lastly, the current system is still a prototype; for proper integration into a UAV-mountable platform, the system should be redesigned as a more robust and compact sensing system.

Enabling integrated depth, multispectral, and color sensing is essential for UAVs flying in forests. 
Our system replaces and improves on the data currently collected from the multiple conventional imaging sensors needed for robot navigation and perception. 
This can enable flying robots to access more environments, fly faster, safer, and collect more useful data.

\section*{Acknowledgments}
The authors would like to thank Yu Han for providing equipment and helping conduct the lab spectroscopy experiments, Dr. Petra D‘Odorico for feedback and loaning the multispectral camera, Alexander Barden for assistance with setting up experiments, and Zoo Zürich for access to the Masoala Hall for data capture.

\bibliographystyle{bibtex/IEEEtran}
\bibliography{bibtex/CG-references,all,bibtex/control}

% Generated by IEEEtran.bst, version: 1.14 (2015/08/26)
\begin{thebibliography}{10}
\providecommand{\url}[1]{#1}
\csname url@samestyle\endcsname
\providecommand{\newblock}{\relax}
\providecommand{\bibinfo}[2]{#2}
\providecommand{\BIBentrySTDinterwordspacing}{\spaceskip=0pt\relax}
\providecommand{\BIBentryALTinterwordstretchfactor}{4}
\providecommand{\BIBentryALTinterwordspacing}{\spaceskip=\fontdimen2\font plus
\BIBentryALTinterwordstretchfactor\fontdimen3\font minus
  \fontdimen4\font\relax}
\providecommand{\BIBforeignlanguage}[2]{{%
\expandafter\ifx\csname l@#1\endcsname\relax
\typeout{** WARNING: IEEEtran.bst: No hyphenation pattern has been}%
\typeout{** loaded for the language `#1'. Using the pattern for}%
\typeout{** the default language instead.}%
\else
\language=\csname l@#1\endcsname
\fi
#2}}
\providecommand{\BIBdecl}{\relax}
\BIBdecl

\bibitem{Loquercio2021}
A.~Loquercio \emph{et~al.}, ``{Learning high-speed flight in the wild},''
  \emph{Science Robotics}, vol.~6, no.~59, p. 5810, 10 2021.

\bibitem{Ren2025}
Y.~Ren \emph{et~al.}, ``{Safety-assured high-speed navigation for MAVs},''
  \emph{Science Robotics}, vol.~10, no.~98, p. 6187, 1 2025.

\bibitem{Geckeler2025d}
C.~Geckeler \emph{et~al.}, ``{Field Deployment of BiodivX Drones in the Amazon
  Rainforest for Biodiversity Monitoring},'' \emph{IEEE Transactions on Field
  Robotics}, vol.~2, no. June, pp. 336--352, 2025.

\bibitem{Charron2020}
G.~Charron \emph{et~al.}, ``{The DeLeaves: a UAV device for efficient tree
  canopy sampling},'' \emph{Journal of Unmanned Vehicle Systems}, vol.~8,
  no.~3, pp. 245--264, 9 2020.

\bibitem{Schedl2021}
D.~C. Schedl \emph{et~al.}, ``{An autonomous drone for search and rescue in
  forests using airborne optical sectioning},'' \emph{Science Robotics},
  vol.~6, no.~55, pp. 1--11, 6 2021.

\bibitem{Geckeler2022}
C.~Geckeler and S.~Mintchev, ``Bistable helical origami gripper for sensor
  placement on branches,'' \emph{Advanced Intelligent Systems}, p. 2200087, 8
  2022.

\bibitem{Geckeler2023b}
C.~Geckeler \emph{et~al.}, ``{Biodegradable Origami Gripper Actuated with
  Gelatin Hydrogel for Aerial Sensor Attachment to Tree Branches},'' in
  \emph{Proceedings - IEEE International Conference on Robotics and
  Automation}, vol. 2023-May.\hskip 1em plus 0.5em minus 0.4em\relax IEEE, 5
  2023, pp. 5324--5330.

\bibitem{Aucone2023a}
E.~Aucone \emph{et~al.}, ``{Drone-assisted collection of environmental DNA from
  tree branches for biodiversity monitoring},'' \emph{Science Robotics},
  vol.~8, no.~74, p. eadd5762, 1 2023.

\bibitem{Kirchgeorg2024}
S.~Kirchgeorg \emph{et~al.}, ``{eProbe: Sampling of Environmental DNA within
  Tree Canopies with Drones},'' \emph{Environmental Science {\&} Technology}, 9
  2024.

\bibitem{Aucone2024}
E.~Aucone \emph{et~al.}, ``{Synergistic morphology and feedback control for
  traversal of unknown compliant obstacles with aerial robots},'' \emph{Nature
  Communications}, vol.~15, no.~1, p. 2646, 3 2024.

\bibitem{Geckeler2024}
C.~Geckeler \emph{et~al.}, ``{Learning Occluded Branch Depth Maps in Forest
  Environments Using RGB-D Images},'' \emph{IEEE Robotics and Automation
  Letters}, vol.~9, no.~3, pp. 2439--2446, 3 2024.

\bibitem{Xu2020a}
Z.~Xu \emph{et~al.}, ``{Tree species classification using UAS-based digital
  aerial photogrammetry point clouds and multispectral imageries in subtropical
  natural forests},'' \emph{International Journal of Applied Earth Observation
  and Geoinformation}, vol.~92, p. 102173, 10 2020.

\bibitem{Jarocinska2023}
A.~Jarocinska \emph{et~al.}, ``{The utility of airborne hyperspectral and
  satellite multispectral images in identifying Natura 2000 non-forest habitats
  for conservation purposes},'' \emph{Scientific Reports}, vol.~13, no.~1, p.
  4549, 3 2023.

\bibitem{MEMSFabryFerot}
J.~Antila \emph{et~al.}, ``{Spectral imaging device based on a tuneable MEMS
  Fabry-Perot interferometer},'' in \emph{Next-Generation Spectroscopic
  Technologies V}, M.~A. Druy and R.~A. Crocombe, Eds., vol. 8374,
  International Society for Optics and Photonics.\hskip 1em plus 0.5em minus
  0.4em\relax SPIE, 2012, p. 83740F.

\bibitem{Gupta2008}
N.~Gupta, ``{Hyperspectral imager development at Army Research Laboratory},''
  in \emph{Infrared Technology and Applications XXXIV}, B.~F. Andresen
  \emph{et~al.}, Eds., vol. 6940, International Society for Optics and
  Photonics.\hskip 1em plus 0.5em minus 0.4em\relax SPIE, 2008, p. 69401P.

\bibitem{poger2001multispectral}
S.~Poger and E.~Angelopoulou, ``Multispectral sensors in computer vision,''
  \emph{Stevens Institute of Technology Technical Report CS Report}, vol.~3,
  2001.

\bibitem{descour1996throughput}
M.~R. Descour, ``Throughput advantage in imaging fourier-transform
  spectrometers,'' in \emph{Imaging spectrometry II}, vol. 2819.\hskip 1em plus
  0.5em minus 0.4em\relax SPIE, 1996, pp. 285--290.

\bibitem{Mooney1995}
J.~M. Mooney, ``{Angularly multiplexed spectral imager},'' in \emph{Imaging
  Spectrometry}, M.~R. Descour \emph{et~al.}, Eds., vol. 2480, International
  Society for Optics and Photonics.\hskip 1em plus 0.5em minus 0.4em\relax
  SPIE, 1995, pp. 65 -- 77.

\bibitem{Gehm2007}
M.~E. Gehm \emph{et~al.}, ``Single-shot compressive spectral imaging with a
  dual-disperser architecture,'' \emph{Opt. Express}, vol.~15, no.~21, pp.
  14\,013--14\,027, Oct 2007.

\bibitem{Wang17PAMI}
L.~Wang \emph{et~al.}, ``Adaptive nonlocal sparse representation for
  dual-camera compressive hyperspectral imaging,'' \emph{IEEE Transactions on
  Pattern Analysis and Machine Intelligence}, vol.~39, no.~10, pp. 2104--2111,
  2017.

\bibitem{Rebecq2018}
H.~Rebecq \emph{et~al.}, ``Emvs: Event-based multi-view stereo—3d
  reconstruction with an event camera in real-time,'' \emph{International
  Journal of Computer Vision}, vol. 126, pp. 1394--1414, 12 2018.

\bibitem{Scheerlinck2019}
C.~Scheerlinck \emph{et~al.}, ``{CED: Color Event Camera Dataset},'' in
  \emph{2019 IEEE/CVF Conference on Computer Vision and Pattern Recognition
  Workshops (CVPRW)}, vol. 2019-June.\hskip 1em plus 0.5em minus 0.4em\relax
  IEEE, 6 2019, pp. 1684--1693.

\bibitem{Marcireau2018}
A.~Marcireau \emph{et~al.}, ``{Event-Based Color Segmentation With a High
  Dynamic Range Sensor},'' \emph{Frontiers in Neuroscience}, vol.~12, no. APR,
  p. 317614, 4 2018.

\bibitem{Himmi2024}
S.~Himmi \emph{et~al.}, ``{MS-EVS: Multispectral event-based vision for deep
  learning based face detection},'' in \emph{2024 IEEE/CVF Winter Conference on
  Applications of Computer Vision (WACV)}.\hskip 1em plus 0.5em minus
  0.4em\relax IEEE, 1 2024, pp. 605--614.

\bibitem{Cao2024}
R.~Cao \emph{et~al.}, ``{Noise2Image: noise-enabled static scene recovery for
  event cameras},'' \emph{Optica}, vol.~12, no.~1, p.~46, 1 2025.

\bibitem{Matsuda15ICCP}
N.~Matsuda \emph{et~al.}, ``Mc3d: Motion contrast 3d scanning,'' in \emph{2015
  IEEE International Conference on Computational Photography (ICCP)}, 2015, pp.
  1--10.

\bibitem{ESL}
M.~Muglikar \emph{et~al.}, ``Esl: Event-based structured light,'' in \emph{2021
  International Conference on 3D Vision (3DV)}, 2021, pp. 1165--1174.

\bibitem{Muglikar2021a}
------, ``{Event Guided Depth Sensing},'' in \emph{2021 International
  Conference on 3D Vision (3DV)}.\hskip 1em plus 0.5em minus 0.4em\relax IEEE,
  12 2021, pp. 385--393.

\bibitem{Ehsan2022}
S.~Ehsan \emph{et~al.}, ``{Event-based RGB-D sensing with structured light},''
  \emph{WACV}, 7 2022.

\bibitem{Arja2025}
S.~Arja \emph{et~al.}, ``{Seeing like a Cephalopod: Colour Vision with a
  Monochrome Event Camera},'' in \emph{2025 IEEE/CVF Conference on Computer
  Vision and Pattern Recognition Workshops (CVPRW)}.\hskip 1em plus 0.5em minus
  0.4em\relax IEEE, 6 2025, pp. 4984--4993.

\bibitem{Yu2025}
B.~Yu \emph{et~al.}, ``{Active Hyperspectral Imaging Using an Event Camera},''
  in \emph{2025 IEEE/CVF Conference on Computer Vision and Pattern Recognition
  (CVPR)}.\hskip 1em plus 0.5em minus 0.4em\relax IEEE, 6 2025, pp. 929--939.

\bibitem{Shah2024a}
S.~Shah \emph{et~al.}, ``{CodedEvents: Optimal Point-Spread-Function
  Engineering for 3D-Tracking with Event Cameras},'' in \emph{2024 IEEE/CVF
  Conference on Computer Vision and Pattern Recognition (CVPR)}.\hskip 1em plus
  0.5em minus 0.4em\relax IEEE, 6 2024, pp. 25\,265--25\,275.

\bibitem{Habuchi2024}
S.~Habuchi \emph{et~al.}, ``{Time-Efficient Light-Field Acquisition Using Coded
  Aperture and Events},'' in \emph{2024 IEEE/CVF Conference on Computer Vision
  and Pattern Recognition (CVPR)}.\hskip 1em plus 0.5em minus 0.4em\relax IEEE,
  6 2024, pp. 24\,923--24\,933.

\bibitem{Alonso2019}
I.~Alonso and A.~C. Murillo, ``{EV-SegNet: Semantic Segmentation for
  Event-Based Cameras},'' in \emph{2019 IEEE/CVF Conference on Computer Vision
  and Pattern Recognition Workshops (CVPRW)}.\hskip 1em plus 0.5em minus
  0.4em\relax IEEE, 6 2019, pp. 1624--1633.

\bibitem{Baek2021}
S.-H. Baek \emph{et~al.}, ``{Single-shot Hyperspectral-Depth Imaging with
  Learned Diffractive Optics},'' in \emph{2021 IEEE/CVF International
  Conference on Computer Vision (ICCV)}.\hskip 1em plus 0.5em minus 0.4em\relax
  IEEE, 10 2021, pp. 2631--2640.

\bibitem{Muglikar2021}
M.~Muglikar \emph{et~al.}, ``{ESL: Event-based Structured Light},'' in
  \emph{2021 International Conference on 3D Vision (3DV)}.\hskip 1em plus 0.5em
  minus 0.4em\relax IEEE, 12 2021, pp. 1165--1174.

\bibitem{Lichtsteiner2008}
P.~Lichtsteiner \emph{et~al.}, ``A 128 x 128 120 db 15 $\mu$s latency
  asynchronous temporal contrast vision sensor,'' \emph{IEEE Journal of
  Solid-State Circuits}, vol.~43, pp. 566--576, 2 2008.

\bibitem{Suh2020}
Y.~Suh \emph{et~al.}, ``{A 1280×960 Dynamic Vision Sensor with a 4.95-{$\mu$}m
  Pixel Pitch and Motion Artifact Minimization},'' in \emph{2020 IEEE
  International Symposium on Circuits and Systems (ISCAS)}, vol.
  2020-Octob.\hskip 1em plus 0.5em minus 0.4em\relax IEEE, 10 2020, pp. 1--5.

\bibitem{Finateu20isscc}
T.~Finateu \emph{et~al.}, ``A 1280x720 back-illuminated stacked temporal
  contrast event-based vision sensor with 4.86$\mu$m pixels, 1.066geps readout,
  programmable event-rate controller and compressive data-formatting
  pipeline,'' in \emph{isscc}, 2020.

\bibitem{Posch11ssc}
C.~Posch \emph{et~al.}, ``A {QVGA} 143 {dB} dynamic range frame-free {PWM}
  image sensor with lossless pixel-level video compression and time-domain
  {CDS},'' \emph{ssc}, vol.~46, no.~1, pp. 259--275, Jan. 2011.

\bibitem{Gallego2019}
G.~Gallego \emph{et~al.}, ``{Event-Based Vision: A Survey},'' \emph{IEEE
  Transactions on Pattern Analysis and Machine Intelligence}, vol.~44, no.~1,
  pp. 154--180, 1 2022.

\bibitem{DVSBiases2023}
R.~Graça \emph{et~al.}, ``Shining light on the dvs pixel: A tutorial and
  discussion about biasing and optimization,'' 2023. [Online]. Available:
  \url{http://arxiv.org/abs/2304.04706}

\bibitem{pmdtec}
P.~P. flexx2 3D~camera, \url{https://3d.pmdtec.com/en/3d-cameras/flexx2/}.

\bibitem{IntelRealSense}
I.~R.~D. cameras, \url{https://www.intelrealsense.com/coded-light/}.

\bibitem{RGBD}
S.~E.~M. Bajestani and G.~Beltrame, ``Event-based rgb sensing with structured
  light,'' in \emph{2023 IEEE/CVF Winter Conference on Applications of Computer
  Vision (WACV)}, 2023, pp. 5447--5456.

\bibitem{MSCOCO}
T.-Y. Lin \emph{et~al.}, ``Microsoft coco: Common objects in context,'' in
  \emph{Computer Vision -- ECCV 2014}, D.~Fleet \emph{et~al.}, Eds.\hskip 1em
  plus 0.5em minus 0.4em\relax Cham: Springer International Publishing, 2014,
  pp. 740--755.

\end{thebibliography}

\end{document}